\documentclass[runningheads]{llncs}

\usepackage{eccv}

\usepackage{eccvabbrv}

\usepackage{graphicx}
\usepackage{booktabs}
\usepackage{multirow}
\usepackage{graphicx}
\usepackage{subcaption}
\usepackage[accsupp]{axessibility}  

\usepackage{hyperref}

\usepackage{orcidlink}

\begin{document}

\title{LoT-Pass: Long-term-robust Image Watermarking for Image to Video Generation} 

\titlerunning{LoT-Pass}

\author{Guanjie Wang\inst{1}\orcidlink{0009-0002-5070-2425} \and
Zehua Ma\inst{1}$^{\dagger}$ \and
Han Fang\inst{1}$^{\dagger}$ \and Weiming Zhang\inst{1}}

\authorrunning{G.~Wang et al.}

\institute{University of Science and Technology of China\\
           \email{wangguanjie@mail.ustc.edu.cn}, 
           \email{\{mzh045, fanghan, zhangwm\}@ustc.edu.cn}\\
           $^{\dagger}$ Corresponding author}

\maketitle

\begin{abstract}
    While Image-to-Video models unlock new creative possibilities, they also raise risks of unauthorized adaptation, deepfakes, and scenario manipulation, making robust authentication and traceability solutions urgent for image owners. While image watermarking offers a solution, existing methods expose a critical vulnerability: due to the inherent semantic drift of generated videos, images watermarks survive only in initial frames. This allows attackers to easily evade detection via \textit{temporal cropping}—discarding early frames to obtain watermark-free derivative videos. Consequently, in an era of pervasive I2V models, \textit{long-term robustness} is imperative for image watermarking. Since simply increasing forward embedding strength fails against severe generative divergence, we introduce \textbf{LoT-Pass}, a watermarking framework built on a new insight: only one-directional robustness enhancement is hard to achieve long-term robustness. We propose a two-way strategy: (1) expand the robustness distance by training under simulated temporal evolution in I2V generation, and (2) recover unextractable frames by reversing them through a temporal inversion module. This shift, from merely resisting distortions to actively rewinding the video to a watermark-friendly state, enables LoT-Pass to maintain extractability even under temporal drift. Experiments on mainstream open-source and commercial I2V models demonstrate that LoT-Pass achieves stronger long-term robustness while preserving imperceptibility, offering a new paradigm for image copyright protection in the era of widespread I2V adoption. \href{https://github.com/MrCrims/LoT-Pass-Long-term-robust-Image-Watermarking-for-Image-to-Video-Generation}{Code}
\end{abstract}

\section{Introduction}
\label{sec:intro}

With the rapid advancement of video generation models, image-to-video (I2V) generation techniques have become increasingly widespread (e.g., Stable Video Diffusion \cite{blattmann2023stablevideodiffusionscaling}, Wan \cite{wan2025}, Hunyuan \cite{kong2024hunyuanvideo}). However, this capability also introduces severe risks of malicious exploitation. For instance, copyrighted artworks or photos of public figures can be misappropriated as input conditions to synthesize unauthorized derivative works, deceptive misinformation, or deepfakes \footnote{https://www.bbc.com/news/articles/cy402zx0w53o} \footnote{https://abcnews.go.com/US/video/fake-images-generated-ai-spreading-social-media-compounding-114824660}. Consequently, establishing reliable provenance tracing not just for images themselves, but across the entire lifecycle of media generated from them, has become an urgent regulatory necessity.

\begin{figure}[t]
    \centering
    \includegraphics[width=\linewidth]{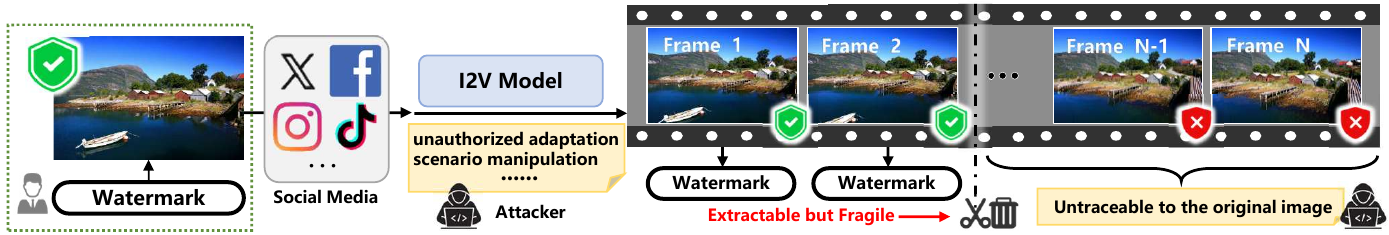}
    \caption{\textbf{IP laundering via temporal cropping in uncontrolled I2V generation.} While source images are protected, attackers exploit the fragile, short-term robustness of existing watermarks during uncontrolled I2V generation. By discarding early extractable frames and isolating late-stage drifted frames, attackers can easily create untraceable derivative works.}
    \label{fig:teaser}
\end{figure}

To mitigate such misuse, recent regulations have called for mechanisms that restrict the generation of sensitive content and support reliable provenance tracing within synthetic media. Digital watermarking, which imperceptibly embeds ownership information into a source image and enables post-hoc verification, is widely regarded as a promising tool for this purpose \cite{hu2025mask, wam, wang2024must, fang2022pimog, hu2025robust, fernandez2024video, ye2023itovefficientlyadaptingdeep,su2025safe,10888991,11299284}.  However, when applied to uncontrolled, black-box I2V generation, existing image watermarks expose a critical security vulnerability. As illustrated in Figure \ref{fig:teaser}, the inherent generative process introduces severe semantic and structural drift over time. While the watermark is reliably detected in the early frames that closely resemble the source image, extraction reliability sharply degrades in later frames as the generated content diverges from the original condition.

This short-term survival creates a trivial bypass for attackers: \textbf{IP Laundering via Temporal Cropping}. By intentionally driving the I2V model to produce temporally extended videos and subsequently discarding the initial frames (which contain both the watermark and obvious visual resemblance to the source), an attacker can easily obtain highly plausible, watermark-free derivative frames. Relying solely on early frames for detection fundamentally fails in this adversarial scenario, as the most valuable and misleading segments of the forged video often lie in the later, heavily drifted stages. Therefore, to ensure persistent provenance, watermarks must maintain \textit{long-term robustness}—surviving deep into the temporal horizon of the generated video.


We argue that achieving such long-term robustness cannot rely exclusively on pushing the limits of forward embedding strength. As the non-linear semantic divergence grows during I2V generation, even the strongest watermarking methods encounter a saturation point beyond which the signal is destroyed. This motivates a fundamental shift in defense strategy: rather than solely forcing the watermark to endure extreme forward drift,  we propose to rewind frames that fall outside this distance back into a watermark-recoverable state.


Based on this insight, we propose \textbf{LoT-Pass}, a bidirectional long-term robust watermarking framework designed for I2V generation. LoT-Pass combats temporal drift through two synergistic modules: (1) a \textit{Temporal Evolution Simulation} module that explicitly models universal generative distortions during training to proactively expand the intrinsic robustness boundary; and (2) a \textit{Temporal Inversion Module} deployed during inference, which approximates backward geometric realignment to pull heavily drifted, late-stage frames back into the decoder’s extractable manifold. 
We evaluate LoT-Pass against SOTA baselines under both open-source and commercial I2V models. Results demonstrate that LoT-Pass achieves significantly improved long-term robustness while preserving imperceptibility. 

    
Our main contributions are summarized as follows:
\begin{itemize}
    \item We identify the critical vulnerability of existing image watermarks under I2V generation, formulating the necessity of \textit{long-term robustness} to defeat IP laundering and temporal cropping attacks in derivative synthetic works.
    \item We propose LoT-Pass, a long-term temporal robust watermarking framework based on an encoder-decoder architecture. By introducing a temporal evolution simulation module during training, LoT-Pass models generate-induced distortions and achieve strong robustness.
    \item We introduce a Temporal Inversion Module at the extraction stage, providing a plug-and-play mechanism to counteract severe temporal degradation and effectively retrieve watermarks from temporally distant frames.
    \item Experiments demonstrate that LoT-Pass outperforms baselines across four representative open-source and also effective on two commercial I2V models.
\end{itemize}


\begin{figure*}[h]
    \begin{center}
        \includegraphics[width=\textwidth]{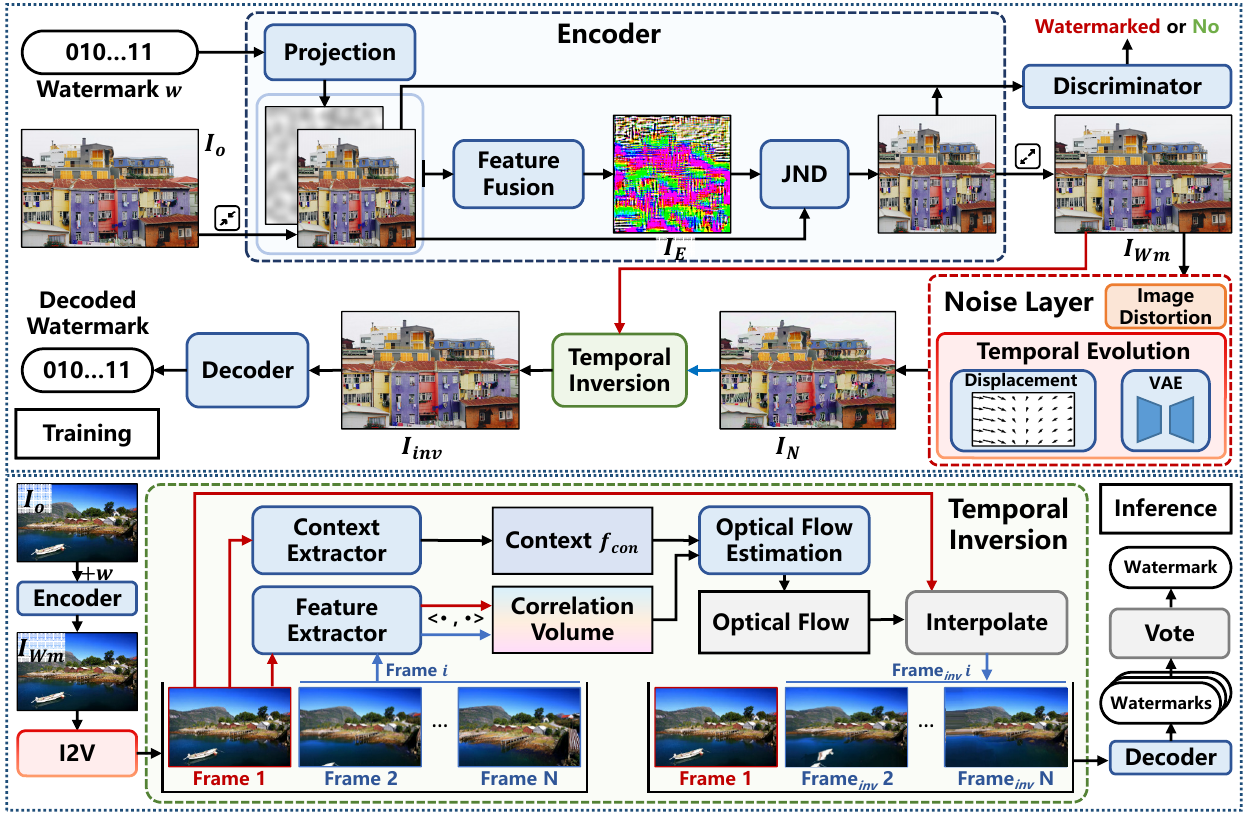}
        \caption{The framework of the proposed LoT-Pass. The encoder embeds the watermark messages into $I_o$ to generate $I_{Wm}$. An adversary discriminator is used to improve the visual quality of $I_{Wm}$. Then, the decoder extracts the watermark message from $I_{Wm}$, which is enhanced by the noise layer. During inference, for an image input, the watermark is directly extracted using the decoder. For a video input, the sequence is first processed by the temporal inversion module, after which each frame is fed into the decoder to extract its corresponding watermark. Finally, all extracted watermarks are aggregated through a voting scheme to obtain the final watermark.}
        \label{fig:pipeline}
    \end{center}
\end{figure*}
\section{Related Work}
\label{sec:related}
\subsection{Image Watermark}
With the rapid advancement of deep learning, numerous studies have explored embedding invisible watermarks into images to establish content provenance. \textit{Ma et al.} \cite{ma2022towards} proposed CIN, a combined reversible and non-reversible mechanism with a non-reversible attention module to address asymmetric extraction under noise attacks. Unlike mainstream architectures, SSLWM \cite{fernandez2022watermarking} embeds messages in the image features extracted by ResNet-50 \cite{he2016deep} and extracts messages using a set of learnable secret keys. MuST \cite{wang2024must} extends the traditional pipeline by incorporating a dedicated localization network to identify all watermarked regions within composite images. Alternatively, WAM \cite{wam} eliminates the need for an explicit localization module by leveraging SAM \cite{kirillov2023segment} 's zero-shot segmentation capabilities, thereby enabling simultaneous region detection and watermark extraction. Furthermore, TrustMark \cite{bui2023trustmark} specifically optimizes the embedding-extraction pipeline to handle diverse input resolutions seamlessly. Recent efforts have also targeted robustness against advanced generative editing. For instance, \textit{Hu et al.} \cite{hu2025robust} introduced Robust-Wide, proposing a novel Partial Instruction-driven Denoising Sampling Guidance to withstand complex, instruction-guided image modifications. Similarly, VINE \cite{lu2024robust} provides an invisible watermarking framework meticulously tailored to resist malicious image editing. Some of these methods can retain limited effectiveness in generated videos, but in most cases, the watermark signal is completely removed in most frames of a generated video.

\subsection{Video Generation Model}

Since Sora \cite{videoworldsimulators2024} demonstrated the potential of Diffusion Transformers (DiTs), video generation has advanced rapidly. Commercial models like Kling \cite{kling} and Hailuo \cite{hailuo} now achieve remarkable long-horizon temporal coherence, which inadvertently exacerbates IP misappropriation risks in extended videos. Concurrently, open-source models have introduced sophisticated architectures for complex spatiotemporal modeling. SVD \cite{blattmann2023stablevideodiffusionscaling} enables stable I2V generation by integrating temporal consistency layers into pre-trained image diffusion models. To efficiently support longer sequences, CogVideoX \cite{yang2024cogvideox} employs a novel 3D causal VAE and an expert Transformer, while Wan \cite{wan2025} incorporates advanced spatiotemporal attention for diverse multimodal conditioning. Furthermore, Hunyuan \cite{kong2024hunyuanvideo} provides a reproducible ecosystem scaling these paradigms. Crucially, while these architectural innovations (e.g., 3D latent compression) dramatically improve video fidelity, they inherently introduce severe non-linear semantic drift, rendering image watermarks fragile over extended frames.

\section{Method}
\label{sec:method}
In this section, we first present the architecture of LoT-Pass, followed by the details of its training and inference.

\subsection{Overall}
As illustrated in Figure \ref{fig:pipeline}, LoT-Pass consists of three trainable models and two auxiliary modules: a watermark encoder $Enc_\theta$, a watermark decoder $Dec_\theta$, a discriminator $Dis_\theta$ used only during training, a JND module for fusing the watermark with the original image, and a noise layer $N$ for adversarial training to enhance robustness. $\theta$ denotes the learnable parameters of the model.

\paragraph{The watermark encoder} $Enc_\theta$ is composed of a message projection module $E_P$ and a feature fusion network $E_{FF}$ based on MUNIT \cite{huang2018munit}. First, the watermark $w \in \{0, 1\}^M$ (with $M$ being the watermark length) is input into the projection module $E_P$, outputting a watermark latent $w_l \in \mathbb{R}^{256 \times 256 \times 3}$. $w_l$ is concatenated with the original image $I_o \in \mathbb{R}^{256 \times 256 \times 3}$ then input into $E_{FF}$ to generate the encoded image $I_E$. Finally, to generate the watermarked image $I_{Wm}$, we utilize the Just-Noticeable-Difference (JND) \cite{475889} module, which is a hand-crafted model of the minimum artifact perceivable by human at every pixel, to fuse $I_E$ and $I_o$:
\begin{align}
    I_{Wm} = I_o + \lambda \times JND(I_E - I_o)
\end{align}
where $\lambda$ is the JND scaling factor to control the strength of the watermark signal.

\paragraph{The discriminator} $Dis_\theta$, implemented based on ResNet \cite{he2016deep}, is designed to distinguish between watermarked and non-watermarked images. During the training of $Enc_\theta$ and $Dec_\theta$, $Dis_\theta$ is incorporated into an adversarial training scheme to constrain the watermarked images generated by $Enc_\theta$ to be closer to the original images, thereby improving imperceptibility. $I_o$ and $I_{Wm}$ are fed into $Dis_\theta$, yielding corresponding outputs $label_o$ and $label_{Wm}$. These outputs $label_o$ and $label_{Wm}$ will be compared against the ground truth to constrain $Dis_\theta$ and $Enc_\theta$'s loss.

\paragraph{Temporal evolution simulation} is designed to simulate the key operation of the I2V generation process during the training of $Enc_\theta$ and $Dec_\theta$, optimizing both models for generating and extracting robust watermark signals. The proposed temporal evolution simulation module includes two additional types of distortions, motivated by the characteristics of video generation, in addition to conventional geometric and non-geometric distortions. First, we employ the VAE of Stable Diffusion \cite{rombach2021highresolution} to reconstruct watermarked images, efficiently simulating the inherent process of video generation, i.e., compression into latent space, noise addition and removal, and decompression into video frames. Second, we propose a random warping strategy to simulate pixel shifts that may occur during video generation. The formulations are as follows:

\begin{equation}
\begin{aligned}
    g &= \text{Upsample}(g_{\text{low}}), \quad g_{\text{low}} \sim \mathcal{N}(0, \sigma^2) \\
    G_{\text{base}}(u,v) =& \left( \frac{2u}{H-1} - 1,\; \frac{2v}{W-1} - 1 \right),  \text{for } u=0,\dots,H-1,\; v=0,\dots,W-1 \\
    I_N &= \text{Interpolate}(I_{Wm},\ G_{\text{base}} + \alpha \cdot g)
\end{aligned}
\end{equation}

We first generate a random two-dimensional grid flow field for low-resolution control points $g_{\text{low}} \in R^{grid_{size}\times grid_{size}\times 2}$ by sampling from a Gaussian distribution with a factor $\sigma$. The low-resolution grid flow field $g_{\text{low}}$ is then upsampled to the target resolution using bicubic interpolation to obtain a smooth and continuous grid flow field $g$. In parallel, we construct a standard normalized sampling grid $G_{\text{base}} \in R^{H\times W \times 2}$ as the reference sampling coordinates without transformation. By adding the normalized displacement field $g$ (with scale factor $\alpha$) to the standard grid, we obtain the final sampling coordinates $G$. Finally, the input image $I_{Wm}$ is resampled according to the final sampling coordinates to produce the randomly distorted image $I_N$. The noise layer $N$ is composed of the temporal evolution simulation module and various common image distortions to enhance the robustness of watermarking algorithms.

\paragraph{The watermark decoder} $Dec_\theta$ is constructed on the ConvNeXt \cite{liu2022convnet} architecture, enabling it to extract the watermark signal from the input image effectively. The noised image $I_N$ is first processed by a ConvNeXt backbone to extract features, which are then passed through an MLP to obtain a soft watermark vector. Finally, the vector is binarized to recover the decoded watermark $w_{dec}$.

\subsection{Training Strategy}
We jointly train $Enc_\theta$, $Dec_\theta$, and $Dis_\theta$ with a two-stage training strategy. In the first stage, the noise layer is not applied, allowing the model to converge to a stable state quickly. In the second stage, the noise layer is activated, and the frequency of each noise is adaptively adjusted at fixed intervals according to the validation. We reweight the different noise according to the following equation:
\begin{align}
w_i &= \frac{\max(e_i, 0.01)}{\sum_{j=1}^{N} \max(e_j, 0.01)}
\end{align}
We require all noise types to participate in the reweighting process, even if the watermarking scheme already achieves an error rate below 1\% under a given noise. This ensures that the model does not discard previously learned robust features while attempting to acquire robustness against other types of noise.
Besides, during training, $I_N$ is fed into the temporal inversion module with probability $P=0.5$. Using $I_{Wm}$ as the reference, the module reconstructs $I_N$ to obtain $I_{inv}$, which is then input into $Dec_\theta$ to recover the watermark. This design enables the model to learn the signal features present in the aligned image.
The overall training objective is formulated as follows:
\begin{align}
    L&oss = \lambda_1L_{enc} + \lambda_2L_{lpips} + \lambda_3L_{adv} + \lambda_4L_{dec} \\
    L&_{enc} =  L_2(I_{Wm} , I_o)
    ,\ L_{adv} = \log(1-Dis_\theta(I_{Wm}))\\
    L&_{lpips} = lpips(I_{Wm}, I_o)
    ,\ L_{dec} = L_2(w_{dec} , w)
\end{align}
where $\lambda_i$ ($i=1,2,3,4$) represent the coefficients for each loss, specifically $\lambda_1 = 1$, $\lambda_2 = 0.1$, $\lambda_3 = 0.1$ and $\lambda_4 = 50$. $L_{lpips}$ is used to constrain the pixel domain's perceptual loss, in which $lpips(\cdot)$ follows \textit{Zhang et al.} \cite{zhang2018unreasonableeffectivenessdeepfeatures}.

\subsection{Inference Strategy}
The inference process of LoT-Pass involves a single mode for $Enc_\theta$ and two modes for $Dec_\theta$, as shown in Figure \ref{fig:pipeline}. For $Enc_\theta$, LoT-Pass supports arbitrary common image resolutions through a residual rescaling and aggregation strategy, which can be formulated as follows:
\begin{align}
\label{eq_res}
    I_{\text{low}} &= \text{DownSample}(I_o) \\
    I_{Wm} &= I_o + \alpha \cdot \text{UpSample}(enc_\theta(I_{\text{low}},w)-I_{\text{low}})
\end{align}
where $\alpha$ indicates the strength factor of the watermark.

For $Dec_\theta$, two modes are provided: an image extraction mode, where the watermark is directly extracted by feeding the image into the decoder, and a video extraction mode, which extends image extraction by incorporating three steps—temporal inversion, decoding, and voting.


\paragraph{Temporal inversion} is based on the optical flow estimation model \cite{teed2020raft}. We first compute the dense optical flow $F$ between the reference frame $I_{\text{ref}}$ (by default, the first frame of the acquired video) and the target frame $I_{\text{tar}}$ using an optical flow model, which represents the displacement of pixels from the reference frame to the target frame. The detailed calculation process is as follows: First, we extract the features $f_{\text{ref}}$ and $f_{\text{tar}}$ of $I_{\text{tar}}$ and $I_{\text{ref}}$.
\begin{equation}
f_{\text{ref}} = \phi(I_{\text{ref}}), \quad 
f_{\text{tar}} = \phi(I_{\text{tar}}), \quad 
f_{\text{con}} = \psi(I_{\text{ref}}) \quad 
\end{equation}
where $\phi(\cdot)$ is the feature extractor and $\psi(\cdot)$ is the context extractor. Then calculate the similarity between pixel pairs.
\begin{equation}
C(i,j,k,l) = \langle f_{\text{ref}}(i,j), f_{\text{tar}}(k,l) \rangle
\end{equation}
where $(i,j)$ denotes the reference frame feature coordinates, $(k,l)$ denotes the target frame feature coordinates, $\langle\cdot,\cdot\rangle$ denotes the inner product. $C$ represents the global correlation volume. 
\begin{equation}
F = \mathcal{H}(\text{Sample}(f_{\text{con}}), C)
\end{equation}
We utilize the optical flow estimation module $\mathcal{H}$ with $C$ and sampled $f_{\text{con}}$ to obtain the final optical flow $F \in \mathbb{R}^{H \times W \times 2}$.

\begin{equation}
\begin{aligned}
X(u,v) &= u + F_x(u,v) , \
Y(u,v) = v + F_y(u,v)
\end{aligned}
\end{equation}
where $(u,v)$ denotes the pixel coordinates of the target frame, and $(F_x, F_y)$ represent the horizontal and vertical components of the optical flow, respectively.
Finally, by applying inverse sampling, we resample the target frame into the reference frame’s coordinate system to obtain the aligned frame.
\begin{equation}
I_{\text{inv}}(u,v) = \text{Interpolate}(I_{\text{tar}}, (X, Y) )
\end{equation}
\paragraph{Voting} is based on a heuristic assumption: \textit{as video frames move further away from the initial frame, the extracted watermarks increasingly resemble samples drawn from a binomial distribution}. For each frame $I_i$, obtain its $I_{\text{inv}}$ and extract the corresponding $w^{(i)}$. The final watermark $w$ can be obtained by a bit-majority vote in the N watermarks $w^{(i)}$.
\begin{equation}
    w^{(1)}, w^{(2)}, ... ,w^{(N)},  w^{(i)} \in \{0,1\}^M
\end{equation}
where $L$ denotes the length of a video. 
\begin{equation}
w_j=
\begin{cases}
1, & \text{if } \sum_{i=1}^L w_j^{(i)} \geq \frac{L}{2}, \\
0, & \text{otherwise}.
\end{cases}
\end{equation}
The final $w$ of the input video is $(w_1, w_2, ... , w_M)$.

\begin{table*}[t]
\small
\begin{center}
\caption{Exhibition on the input image resolution of the model, capacity, bits per pixel (BPP), model parameters, visual metrics (PSNR, SSIM, and LPIPS), and inference cost (inference time and average GPU memory peak usage).}
\resizebox{\textwidth}{!}{
\setlength{\tabcolsep}{.06in}{
\begin{tabular}{c|cccccccccccc}
    \toprule[1.5pt]
    \multirow{2}{*}{Method} & \multicolumn{3}{c}{Input Setting} & \multicolumn{3}{c}{\# Params (M)} & \multicolumn{3}{c}{Visual Metric} & \multicolumn{3}{c}{Inference Cost} \\
    \cmidrule(lr){2-4} \cmidrule(lr){5-7} \cmidrule(lr){8-10} \cmidrule(lr){11-13}
    & Resolution & Capacity (bits) & BPP(\%) $\uparrow$ & Encoder & Decoder & Total & PSNR $\uparrow$ & SSIM $\uparrow$ & LPIPS $\downarrow$ & Encoder Time(s) $\downarrow$ & Decoder Time(s) $\downarrow$ & Memory(GB) $\downarrow$ \\
    SSLWM & 224 & 30 & 0.06 & - & 26.42 & 26.42 & 36.15 & 0.9545 & 0.012 & 2.3 & 0.009 & 0.27 \\
    CIN & 128 & 30 & \textbf{0.18} & 5.29 & 5.00 & 10.30 & 38.80 & \underline{0.9866} & \underline{0.008} & 0.025 & 0.014 & \textbf{0.15} \\
    MuST & 256 & 30 & 0.05 & 0.3 & 0.63 & 0.9 & 36.33 & 0.9518 & 0.013 & 0.081 & \underline{0.007} & 0.36 \\
    TrustMark & 256 & 100 & \underline{0.15} & 8.24 & 22.61 & 30.86 & 38.23 & 0.9819 & \underline{0.008} & \textbf{0.006} & \textbf{0.006} & 0.41 \\
    WAM & 256 & 32 & 0.05 & 1.1 & 96.0 & 97.1 & \underline{39.43} & 0.9793 & \underline{0.007} & \underline{0.015} & 0.015 & 0.51 \\
    VINE & 256 & 100 & \underline{0.15} & 958.2 & 84.58 & 1042.79 & 36.74 & 0.9712 & 0.010 & 0.077 & 0.008 & 4.43 \\
    Robust-Wide & 512 & 64 & 0.02 & 29.73 & 395.48 & 425.21 & \textbf{39.62} & \textbf{0.9940} & \textbf{0.002} & 0.027 & 0.017 & 2.81 \\
    LoT-Pass & 256 & 32 & 0.05 & 3.7 & 47.4 & 51.1 & 38.58 & 0.9840 & 0.010 & \textbf{0.006} & 0.009 & \underline{0.27} \\
    \bottomrule[1.5pt]
\end{tabular}}
}
\label{tab:metric}
\end{center}
\end{table*}

\section{Experiments}
\label{sec:exp}

\subsection{Experiment Settings}
\label{I2VBench}
\paragraph{Datasets.} We adopt the widely used DIV2K \cite{Agustsson_2017_CVPR_Workshops} dataset as the test image set for watermarking methods. To construct the corresponding text prompts, we employ the Qwen2.5-VL-7B-Instruct \cite{Qwen2.5-VL} model to generate descriptions of each image. For I2V models supporting negative prompts, we uniformly apply a negative prompt shown in the Supplementary. In addition, we test the robustness against classic image distortions on the sampled validation set of MS COCO \cite{lin2014microsoft} and Flickr2K \cite{Lim_2017_CVPR_Workshops}, a total of 200 images.

\paragraph{Metrics.} We evaluate the imperceptibility of watermark models using standard metrics, including PSNR \cite{almohammad2010stego}, SSIM \cite{wang2004image}, and LPIPS \cite{zhang2018unreasonableeffectivenessdeepfeatures}. Watermark capacity is measured by bits per pixel (BPP). For watermark extraction, bit accuracy (ACC) serves as the most fundamental measure of extraction performance. To extend this evaluation to the image-to-video (I2V) setting, we further introduce two video-level robustness metrics: frame accuracy ($\boldsymbol{FACC}$) and video accuracy ($\overline{VACC}$). Based on $\boldsymbol{FACC}$, we introduce the concept of \textbf{Robust Diffusion Distance (RDD)} to evaluate the robust persistence of watermark signals temporally under the modifications induced by Image-to-Video models. These metrics can be formally computed as follows.
\begin{align}
    \boldsymbol{FACC} &= (\overline{ACC_{F_1}}, \dots, \overline{ACC_{F_L}}) \\
    \overline{ACC_{F_i}} &= \frac{\sum^T_{j=1}ACC^j_{F_i}}{T} , \quad i=1,\dots,L \\
    RDD &= \arg\min_i \overline{ACC_{F_i}} < \tau, \quad \overline{ACC_{F_i}}\in \boldsymbol{FACC} \\
    \overline{VACC} &= \frac{\sum^N_{i=1}\sum^L_{j=1}ACC^j_{F_i}}{N\times L} \label{eq_vacc}
\end{align}
where $\overline{ACC_{F_i}}$ denotes the average accuracy of $i$-th frame across test videos, $T$ denotes the number of test videos, and $L$ denotes the length of a video. Formally, RDD is defined as the first frame index at which the accuracy drops below the threshold $\tau=0.9$, indicating the temporal boundary of robustness.

\paragraph{Training Implementation.} We train LoT-Pass using the MS COCO dataset \cite{lin2014microsoft} at a resolution of $256\times256$. The whole framework is implemented by PyTorch \cite{paszke2019pytorch} and all experiments executed on a single NVIDIA RTX A6000 GPU. We utilize AdamW \cite{loshchilov2017decoupled} as the optimizer. The watermark messages are randomly generated sequences of $32$ bits. The batch size in the training stage is set to $16$, and the LoT-Pass models are trained for $250$ epochs with an initial learning rate $= 0.0001$. 
Details are in the Supplementary.

\paragraph{Baselines.} We compare LoT-Pass with seven watermark baselines, all utilizing their officially open-source implementations. These baselines include SSLWM \cite{fernandez2022watermarking}, CIN \cite{ma2022towards}, MuST \cite{wang2024must}, VINE-R \cite{lu2024robust}, TrustMark \cite{bui2023trustmark}, Robust-Wide \cite{hu2025robust}, Watermark Anything Model(WAM) \cite{wam}. Although the baselines were trained at different fixed resolutions, we apply watermark residual scaling (using bilinear interpolation methods) to standardize all of them to the original shape of the input images. For fair comparison, all main-text metrics ($\overline{VACC}$ and RDD) are computed before voting; post-voting results are exhibited in the Supplementary. 

\paragraph{Baseline Settings.} In Table \ref{tab:metric}, we provide a comprehensive comparison between LoT-Pass and baseline methods, including input/output sizes, message capacity, perceptual quality metrics of the watermarked images, as well as inference time and computational cost for each method. It is worth noting that, to ensure consistent watermark strength, we adjust the PSNR of all watermarked images to be greater than 36 dB. This prevents unfair comparisons that could arise from sacrificing visual quality in exchange for robustness. Regarding the abnormally long encoding time of SSLWM, this can be explained as follows: SSLWM embeds watermarks by first fine-tuning a set of parameters for each batch of images, followed by data augmentation. When a new batch of images arrives, the parameters must be retrained. We argue that the data augmentation stage alone is insufficient for watermark embedding, and the preceding training stage plays a crucial role in this process. 

\begin{figure*}[t]
    \centering
    \begin{subfigure}{0.48\textwidth}
        \includegraphics[width=\linewidth]{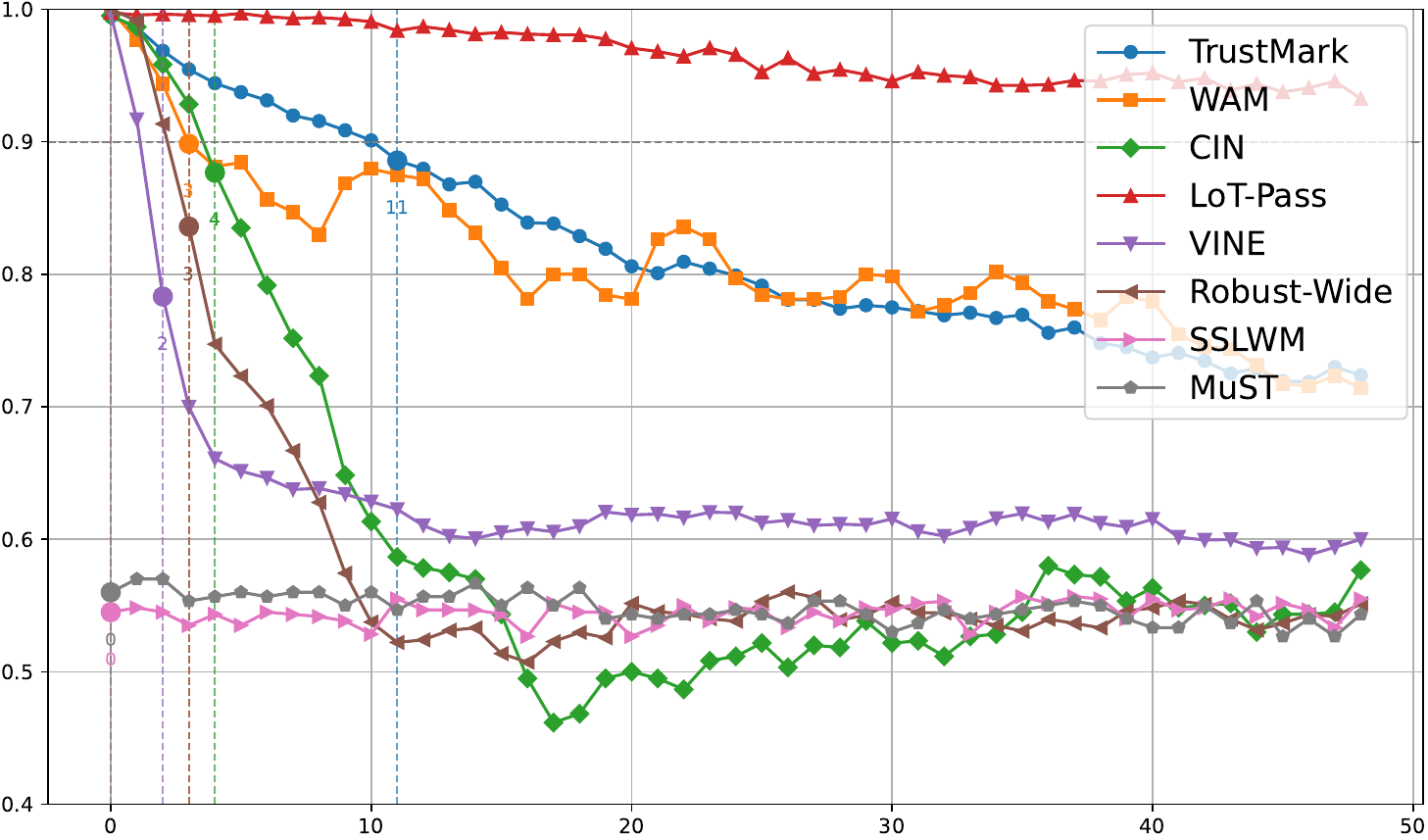}
        \caption{CogVideoX}
    \end{subfigure}
    \begin{subfigure}{0.48\textwidth}
        \includegraphics[width=\linewidth]{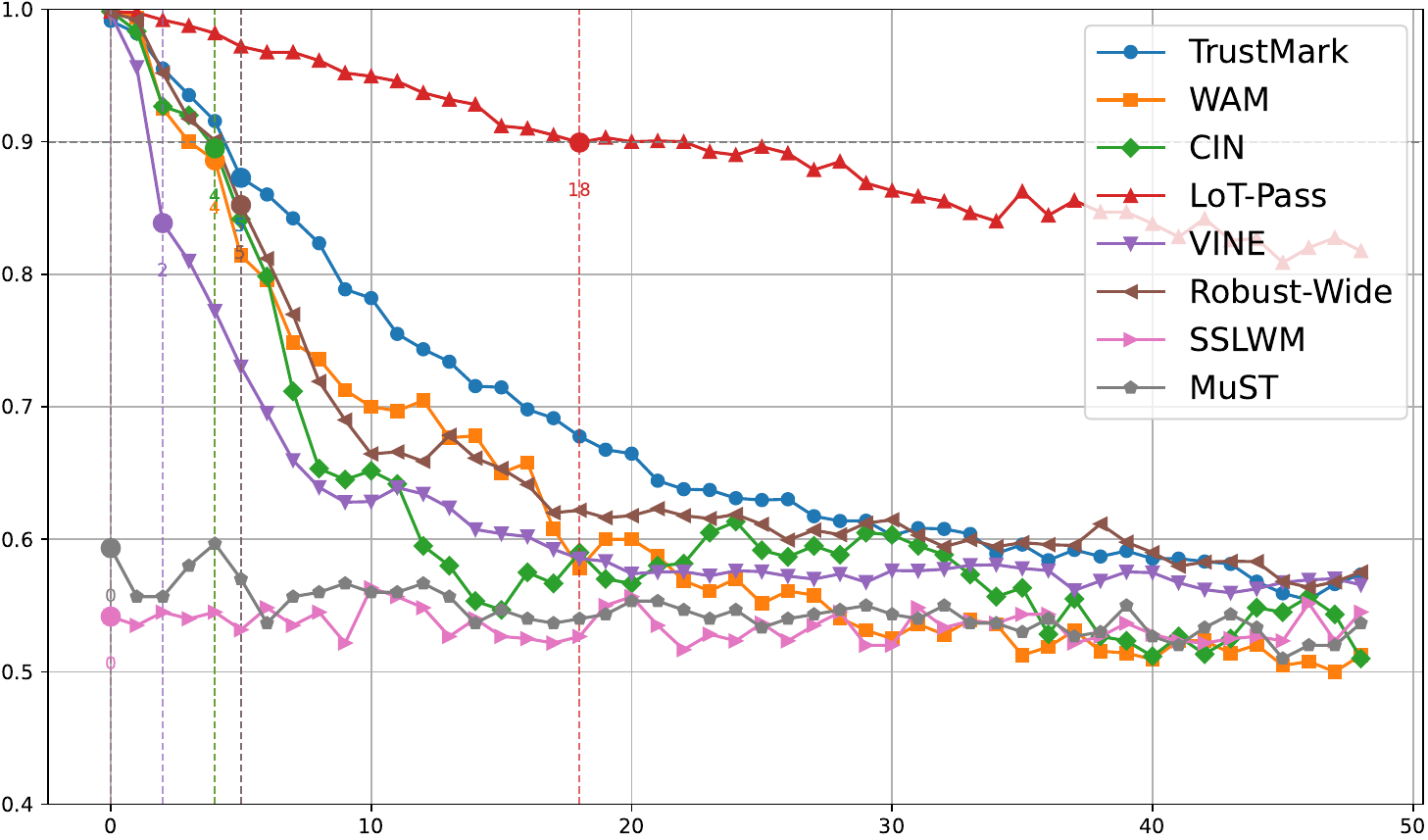}
        \caption{Hunyuan}
    \end{subfigure}

    \begin{subfigure}{0.48\textwidth}
        \includegraphics[width=\linewidth]{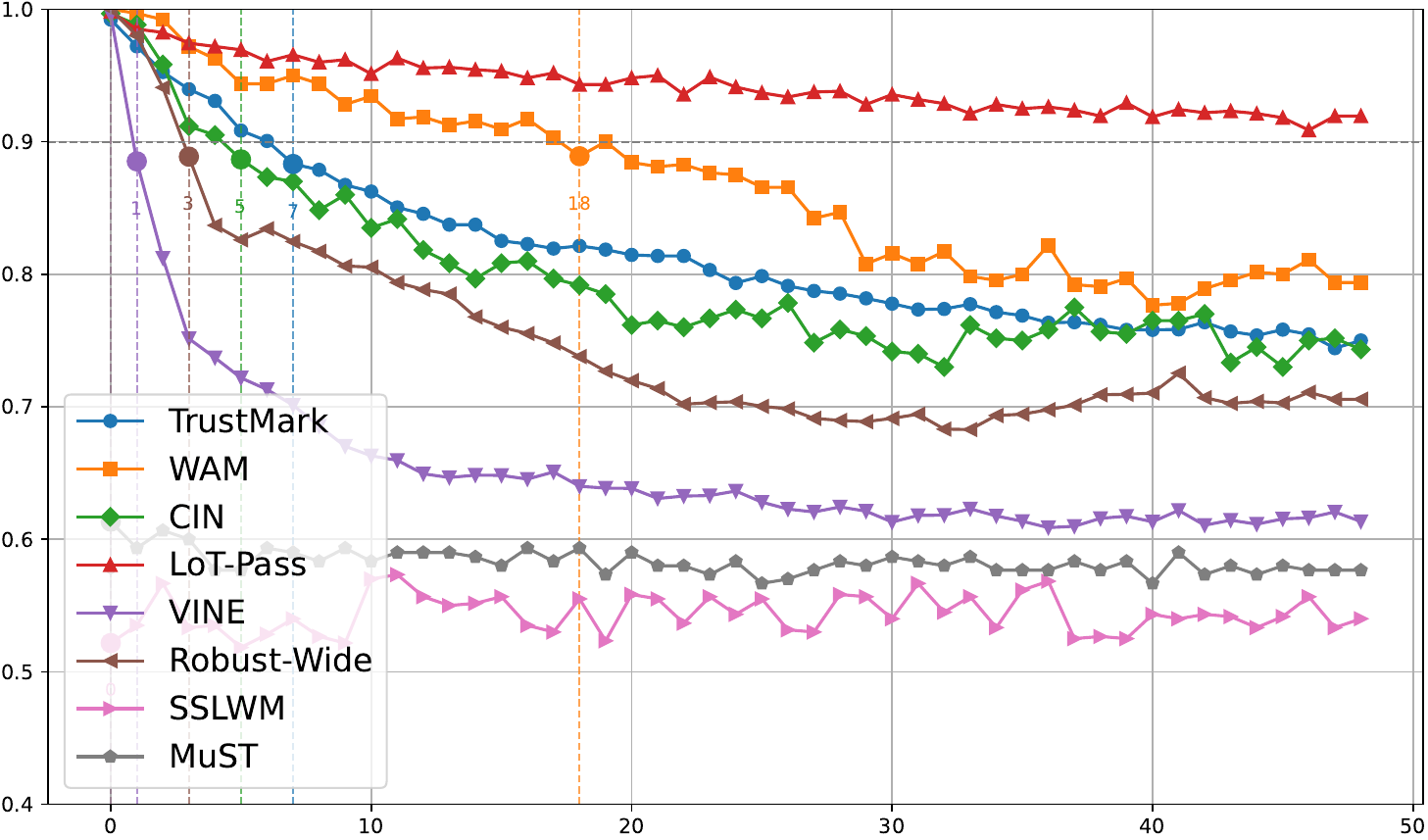}
        \caption{Wan}
    \end{subfigure}
    \begin{subfigure}{0.48\textwidth}
        \includegraphics[width=\linewidth]{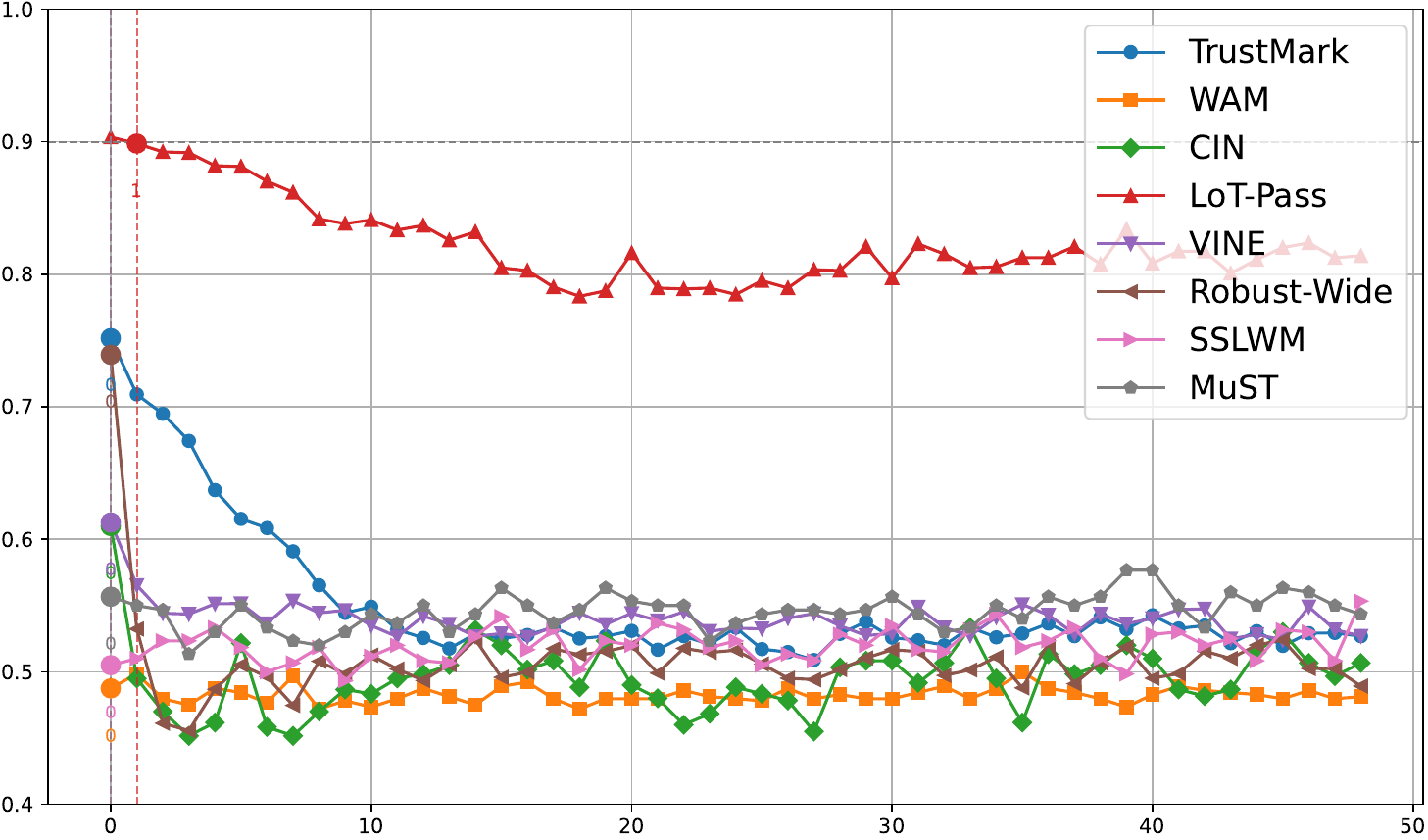}
        \caption{SVD}
    \end{subfigure}

    \caption{Frame accuracy ($\boldsymbol{FACC}$) trends of different watermarking methods across four target models.}
    \label{fig:facc_trends}
\end{figure*}

\begin{figure*}[t]
    \centering
    \begin{subfigure}{0.48\textwidth}
        \includegraphics[width=\linewidth]{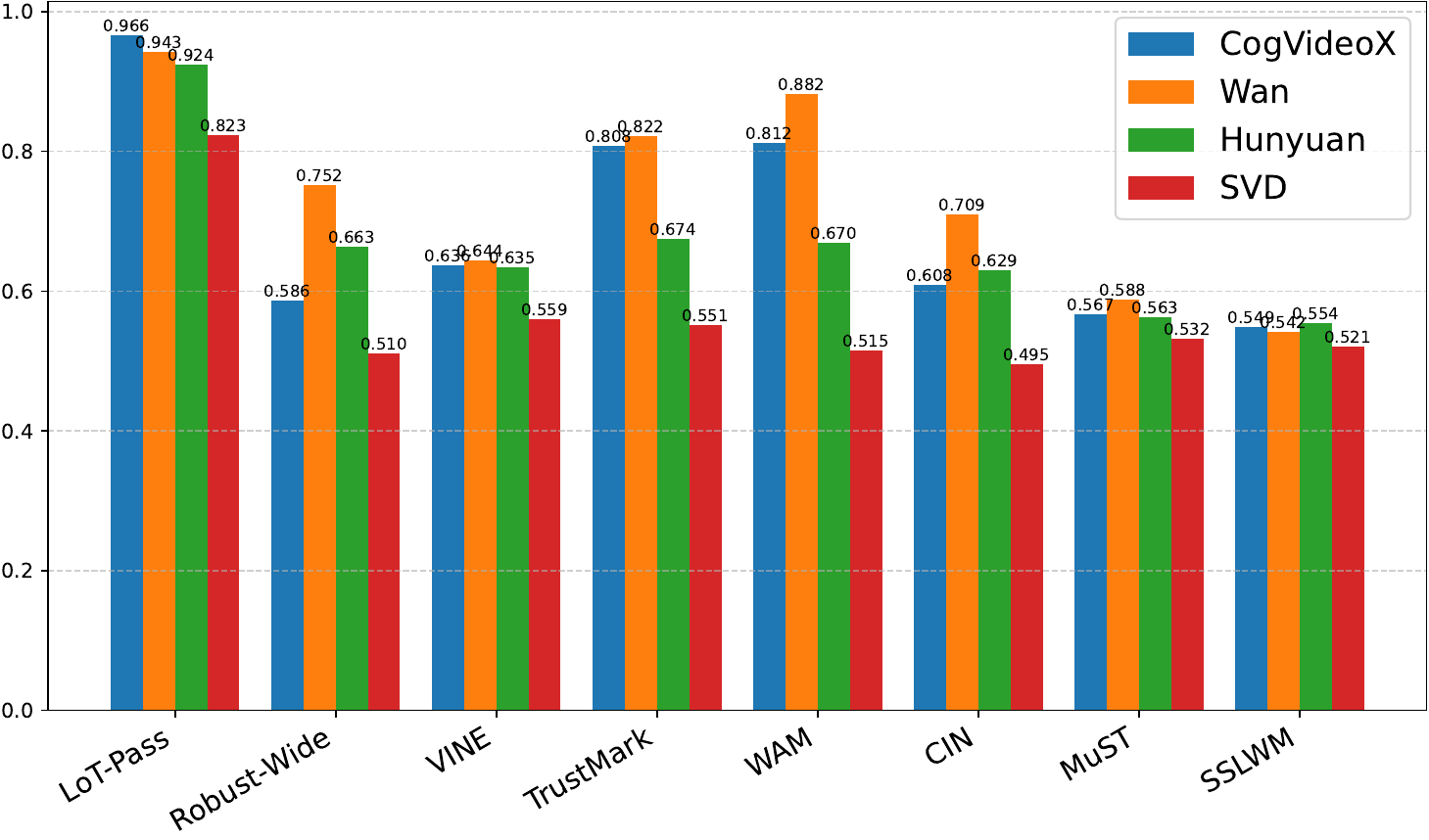}
        \caption{$\overline{VACC}$\ \ Comparison}
    \end{subfigure}
    \begin{subfigure}{0.48\textwidth}
        \includegraphics[width=\linewidth]{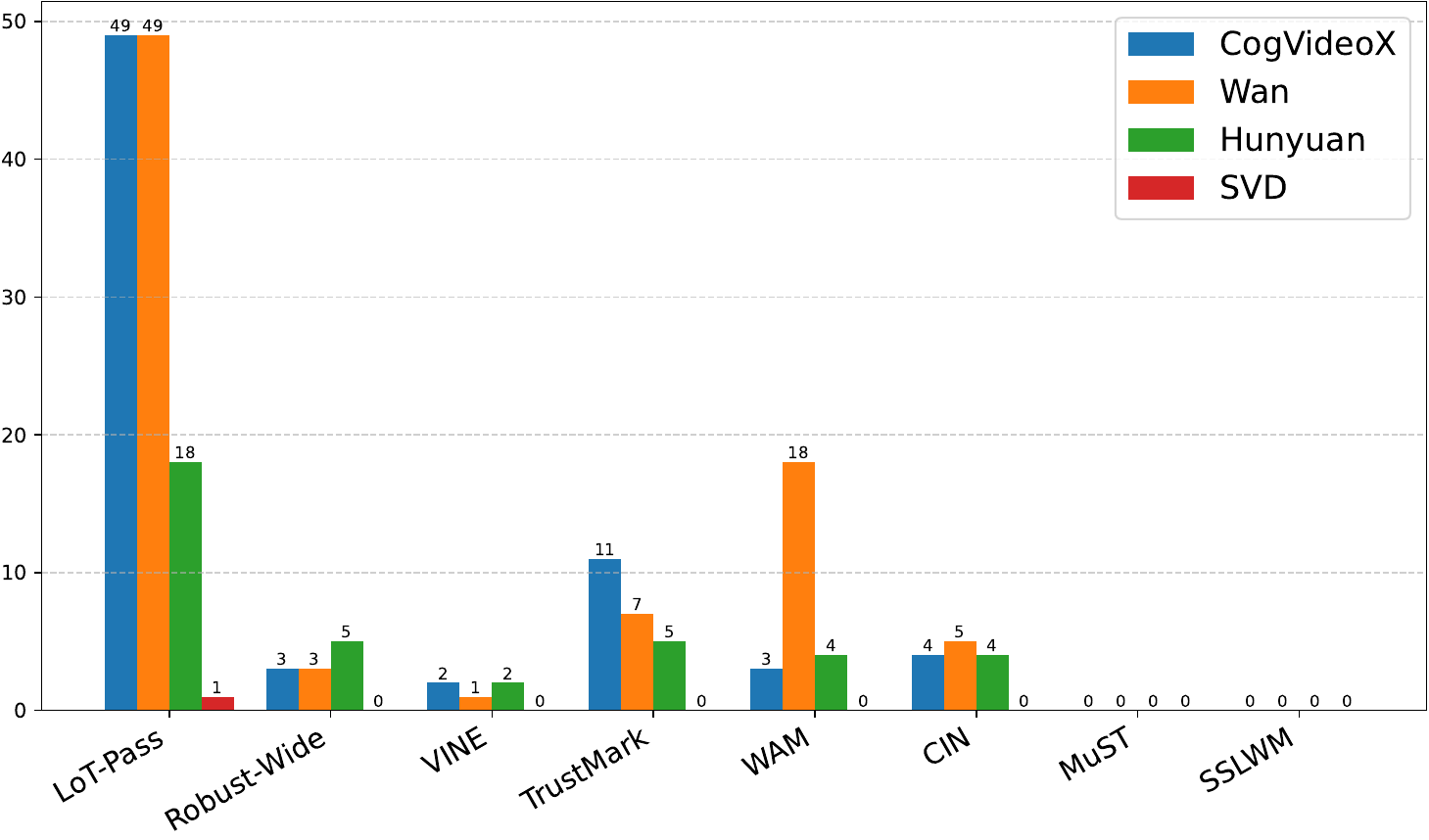}
        \caption{$RDD$\ \ Comparison}
    \end{subfigure}
    \caption{Comparison of video accuracy ($\overline{VACC}$) and robustness diffusion distance ($RDD$) across four target models.}
    \label{fig:summary}
\end{figure*}

\subsection{Results and Analysis}
\label{results}

\subsubsection{Test Open-source I2V Model}

\begin{figure}[t]
    \centering
    \includegraphics[width=0.7\linewidth]{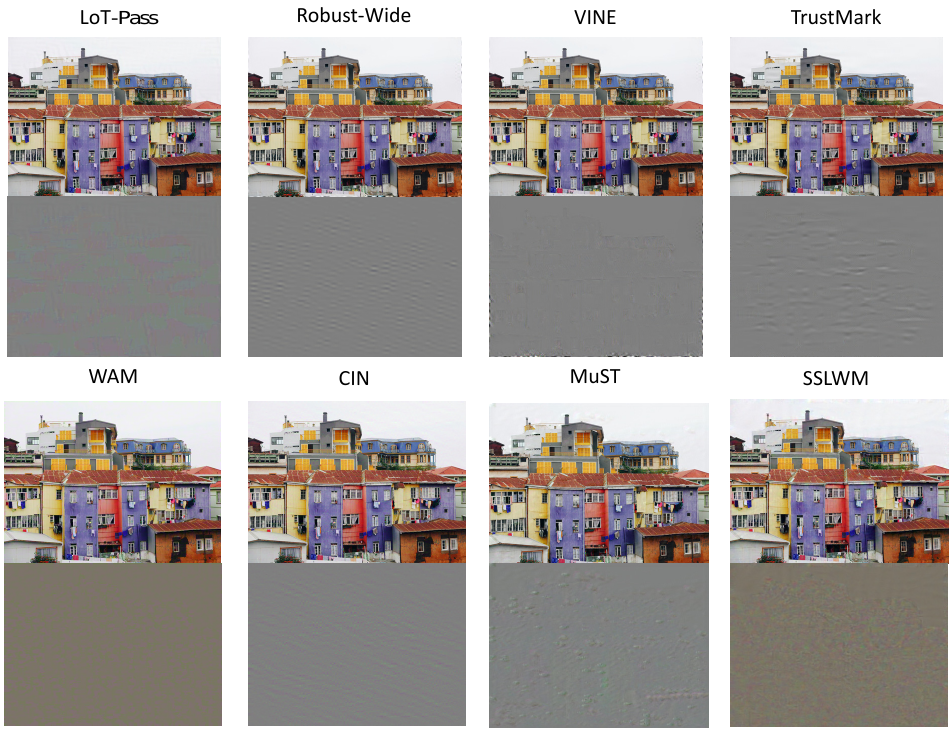}
    \caption{Watermark pattern visualization of each method. $pattern = \alpha\times(I_{Wm}-I_o)$, where $\alpha=3$ to make the pattern more easily recognizable.}
    \label{fig:visualization}
\end{figure}

As shown in Figure \ref{fig:facc_trends} and Figure \ref{fig:summary}, we evaluate LoT-Pass and multiple baselines across four representative open-source image-to-video generation models: CogVideoX \cite{yang2024cogvideox}, Wan2.1 \cite{wan2025}, and Hunyuan \cite{kong2024hunyuanvideo} as \textbf{prompt-guided} I2V models, and Stable Video Diffusion (SVD) \cite{blattmann2023stablevideodiffusionscaling} as a \textbf{non-prompt-guided} I2V model, each generating 50 480P videos for watermark verification. Detailed settings are in the Supplementary. Results demonstrate that LoT-Pass (triangular red line) achieves SOTA performance across all models.
Although WAM and TrustMark were not trained with generative model components such as VAEs, they still exhibit stronger robustness than other baselines. The watermark patterns of different methods are visualized in Figure \ref{fig:visualization}. Observation reveals that the watermark patterns of LoT-Pass, WAM, and TrustMark all exhibit distinct large-scale speckled patterns. As shown in Table \ref{tab:latent_diff}, we compute the L2 distance between watermarked and unwatermarked images in the SD-v1.4 VAE latent space, where LoT-Pass, TrustMark, and WAM achieve higher values than other methods. We believe that such a speckled distribution allows watermark signals to remain more distinguishable after VAE compression, leading to greater retention during the generation process and, consequently, a larger $RDD$. 
Nevertheless, as the generation progresses, the influence of the input image on subsequent latents gradually weakens, leading all watermarking methods to exhibit a declining trend in $\boldsymbol{FACC}$ with increasing temporal distance from the initial frame. Furthermore, for frames that exhibit little to no visual similarity to the original image, we believe that their copyrights and dissemination should likewise be considered independent of the original image. Moreover, under SVD, most methods yield $RDD$ values of zero. We argue this is due to SVD’s lack of prompt conditioning, where video generation relies solely on the input image, resulting in uncontrolled degradation relative to the input image. 

\begin{table}[t]
    \small
    \begin{center}
    \caption{VAE latent difference across I2V robust methods. }
    \resizebox{0.9\linewidth}{!}
{   \begin{tabular}{c|cccccc}
    \toprule[1.5pt]
        Methods &  LoT-Pass & Robust-Wide & VINE & TrustMark & WAM & CIN\\
        \midrule
         Latent Diff.& 0.0023 & 0.0009 & 0.0005 & 0.0015 & 0.0018 & 0.0010\\
         \bottomrule[1.5pt]
    \end{tabular}}
    
    \label{tab:latent_diff}
    \end{center}
\end{table}

\begin{table*}[t]
\small
\begin{center}
\caption{The average $\overline{ACC}(\%)$ results under distortion settings. The parameters for each test image are randomly sampled from the range defined by the upper and lower bounds (represented by the two lines in the settings). \textbf{Bold} results indicate the best outcome among the comparisons, while \underline{underlined} results signify the second-best. More results and details are in the Supplementary.}
\resizebox{\textwidth}{!}
{ 
\begin{tabular}{c|cccccccccccc}
    \toprule[1.5pt]
    Type & Identity & RealCrop & Crop  & RemoveLines & Resize & JPEG & GB & GN & Brightness & Contrast & Saturation & R.Bending  \\
    \midrule
    \multirow{2}{*}{Settings} & - & 40\% &40\% &$ freq_x=2$ & 0.3 & $Q=50$ & $\sigma=0.5$ & $s=0.25$ & 0.5 & 0.5 & 0.5 & $\sigma=0.5$ \\
    & - & 80\% & 80\% & $freq_y=2$ & 1.3 & $Q=80$ & $\sigma=1.5$ & $s=0.75$ & 2 & 2 & 2 & $\sigma=0.5$ \\
    \toprule[1.5pt]
     SSLWM & 98.22 & 50.16  & 73.48 & 94.27 &  94.50 & 89.18 & 96.18 & 92.09 & 84.90 & 76.58 & 83.60 & 56.82 \\
    CIN & 99.67 & 50.23 & 90.98 & 98.68 & 99.50 & 82.36 & 98.47 & 96.70 & \underline{98.63} & \textbf{99.90} & 99.03 & 59.47 \\
    MuST & 96.73 & 88.97 & 83.90 & 93.30 & 92.37 & 73.03 & 94.37 & 65.93 & 75.70 & 79.87 & 84.34 & 65.00 \\
    TrustMark & 99.67 & 65.26 & 92.13 & 98.45 & 98.79 & 98.75 & 99.18 & 88.82 & 92.99 & 94.68 & 99.06 & 59.26 \\
    WAM & 99.22 & \textbf{99.09} & \underline{95.59} & 99.72 & 96.00 & 80.91 & 99.50 & \underline{99.31} & 94.69 & 95.78 & 97.00 & \underline{74.78} \\
    VINE & 99.57 & 50.38 & 51.82 & 99.39 & 99.13 & 96.03 & 99.52 & 97.00 & 97.39 & 99.01 & 99.38 & 56.65 \\
    Robust-Wide & \textbf{99.99} & 50.82 & 89.03 & \textbf{99.99} & \textbf{99.99} & \underline{98.90} & \textbf{99.99} & 98.76 & 98.18 & \underline{99.39} & \textbf{99.99} & 60.88 \\
    LoT-Pass & \underline{99.98} & \underline{98.09} & \textbf{97.38} & \underline{99.91} & \underline{99.91} & \textbf{99.72} & \textbf{99.94} & \textbf{99.75} & \textbf{99.56} & 99.06 & \underline{99.91} & \textbf{97.53} \\
    \bottomrule[1.5pt]
\end{tabular}
}
\label{tab:robust_classic}
\end{center}
\end{table*}

\subsubsection{Test Commercial Model}
\begin{figure}[t]
    \centering
    \includegraphics[width=0.7\linewidth,height=0.18\linewidth]{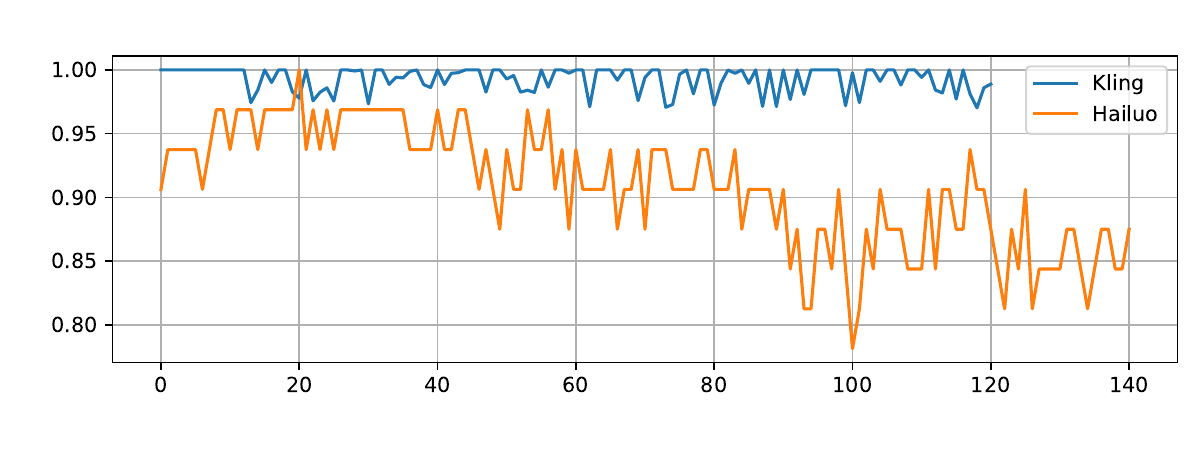}
    \caption{$\boldsymbol{FACC}$ of Kling and Hailuo.}
    \label{fig:commercial}
\end{figure}
In addition to the four open-source models discussed above, we further evaluate LoT-Pass on two commercial models, including Kling Video 2.1 \cite{kling} and Minimax Hailuo 02 \cite{hailuo}, generating ten videos each.
The same image-prompt pairs used for the open-source models are employed to ensure comparability.
As shown in Figure \ref{fig:commercial}, compared with the results on open-source models, the performance on commercial models is better. By examining the quality of the generated videos, we find that commercial models generally produce higher-quality and more temporally coherent outputs, which suggests that improved video generation quality facilitates stronger preservation of watermark signals from the source image. Besides, we hypothesize that commercial models may incorporate certain specialized mechanisms that allow the input image to influence even temporally distant frames, thereby leading to consistently strong $RDD$ performance ($RDD_{Kling}=120, RDD_{Hailuo}=47$).

\begin{table}[t]
\small
\begin{center}
\caption{$\overline{VACC}$ and $RDD$ of LoT-Pass under frame drop.}
\resizebox{0.8\linewidth}{!}
{ 
\begin{tabular}{c|ccccc}
    \toprule[1.5pt]
    $\overline{VACC}/RDD$ & 0 & 5\% & 10\% & 15\% & 20\% \\
    \midrule
    CogVideoX & 0.966/49 & 0.953/49 & 0.949/36 & 0.931/20 & 0.910/13\\
    Hunyuan & 0.924/18 & 0.895/14 & 0.874/8 & 0.858/2 & 0.795/0 \\
    Wan & 0.943/49 & 0.928/36 & 0.912/21 & 0.894/15 & 0.874/7 \\
    SVD & 0.823/1 & 0.698/0 & 0.629/0 & 0.625/0 & 0.582/0 \\
    \bottomrule[1.5pt]
\end{tabular}
}
\label{tab:drop}
\end{center}
\end{table}

\subsubsection{Robustness against Frame Drop}
We further evaluate the robustness of LoT-Pass by discarding a certain proportion from the beginning of video frames on the same video used in Figure \ref{fig:facc_trends}. As shown in Table \ref{tab:drop}, LoT-Pass demonstrates superior performance over the other baselines in most cases, even when these early frames, closely resembling the source image, are removed. This indicates that LoT-Pass exhibits inherent robustness to the I2V process regardless of temporal inversion, a conclusion further supported by our ablation studies.

\subsubsection{Test Classic Noise}
We evaluated each method under classical distortion scenarios. As shown in Table \ref{tab:robust_classic}, WAM, Robust-Wide, and LoT-Pass all demonstrated quite satisfactory performance. 
Watermarking methods that excel in the I2V task, such as TrustMark and WAM, also exhibit strong robustness against classical geometric distortions (e.g., cropping). LoT-Pass remains competitive under geometric transformations and demonstrates superior robustness to random bending attacks (random pixel displacements), achieving over 22\% improvement over the second-best method. Meanwhile, it further demonstrates superior robustness to numerical distortions (e.g., Gaussian blur, Gaussian noise) compared to other methods. This observation indicates that the I2V process inherently combines variations in pixel values and motions, requiring robustness across both dimensions for resistance. Moreover, results validate the effectiveness of our noise layer, which jointly accounts for both distortion types. 

\begin{table}[t]
\small
\vspace{-3pt}
\begin{center}
\caption{$\overline{VACC}$ with different motion intensity.}
{ 
\begin{tabular}{c|cccc}
    \toprule[1.5pt]
    Models & CogVideoX & Hunyuan & Wan & SVD \\
    \midrule
    $\overline{VACC}$ / Low Motion Video Num & 0.971/34 & 0.936/15 & 0.956/19 & 0.883/2 \\
    $\overline{VACC}$ / High Motion Video Num & 0.955/16 & 0.919/35 & 0.935/31 & 0.821/48 \\
    \bottomrule[1.5pt]
\end{tabular}
}
\label{tab:motion}
\end{center}
\vspace{-3pt}
\end{table}

\begin{figure}[t]
\vspace{-3pt}
    \centering
    \includegraphics[width=\linewidth]{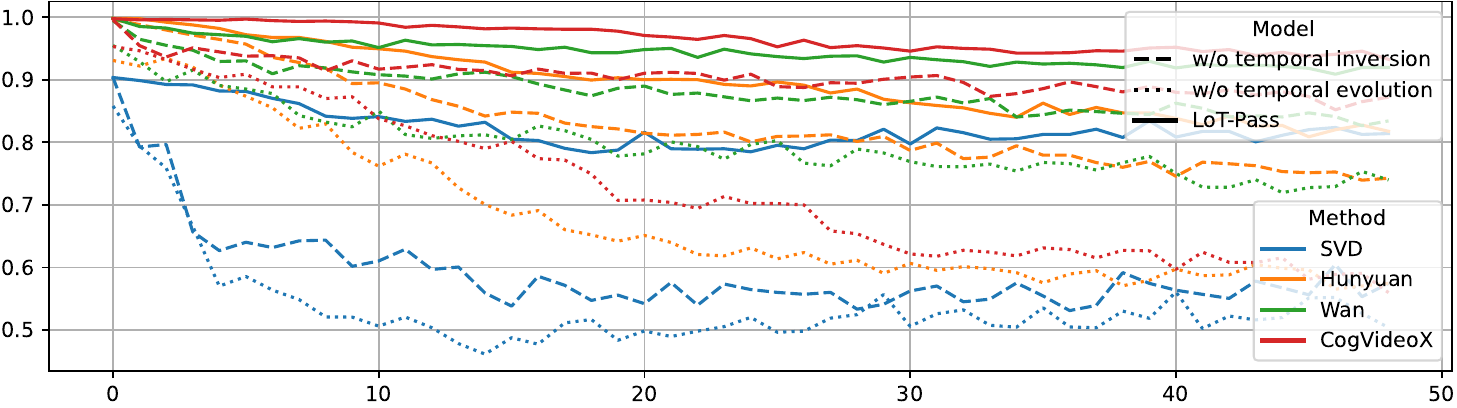}
    \caption{$\boldsymbol{FACC}$ comparison of LoT-Pass, w/o temporal inversion (dashed line), and w/o temporal evolution (dotted line).}
    \vspace{-3pt}
    \label{fig:ablation}
\end{figure}

\subsection{Ablation Study}
\label{ablation}

\paragraph{Effectiveness of Temporal Evolution Simulation.} As shown in Figure \ref{fig:ablation}, we compared two models trained for the same iterations: one with the temporal evolution simulation module in the noise layer and one without (both using temporal inversion during testing). It is evident that models trained without temporal evolution significantly underperform compared to LoT-Pass. This demonstrates that our temporal evolution simulation module effectively captures the key characteristics of I2V generation: VAE compression-reconstruction is an indispensable process, and random displacement simulates pixel motion.

\paragraph{Effectiveness of Temporal Inversion.} To assess the impact of temporal inversion, we evaluated both settings on the same generated videos and computed $\boldsymbol{FACC}$, as shown in Figure \ref{fig:ablation}. Applying the temporal inversion module yields a significant improvement compared to direct extraction from the raw video. 
Temporal inversion remaps unknown temporal perturbations (motion shifts, pixel changes, et al) from video generation models back to the pixel domain; temporal evolution simulates temporal perturbations in the pixel domain during training. The combination of two approaches enables LoT-Pass to achieve strong performance.

\paragraph{Influence of Motion Intensity.}
Using VBench \cite{huang2023vbench}, we categorize videos into low- and high-motion subsets based on temporal flickering, motion smoothness, and dynamic degree. As shown in Table \ref{tab:motion}, although intense generative motions cause more severe spatial drift and slightly degrade the extraction accuracy ($\overline{VACC}$), LoT-Pass maintains exceptional robustness. Even in high-motion scenarios, it achieves a $\overline{VACC}$ exceeding 0.91 on most mainstream models, proving its strong resilience to complex temporal dynamics.

\section{Conclusion}
\label{conclusion}
In this work, we introduce the cross-modal robust watermarking challenge in the image-to-video (I2V) generation for the first time. To systematically evaluate this challenge, we propose three effective evaluation metrics: $RDD$, $\boldsymbol{FACC}$, and $\overline{VACC}$. To address the challenge, we present LoT-Pass, which strengthens watermark signals along the forward temporal axis through a temporal evolution simulation module and enhances backward temporal recovery via the temporal inversion module. Experiments on multiple open-source and commercial video generation models demonstrate that LoT-Pass achieves SOTA performance.

\section*{Acknowledgements}
This work was supported in part by the Natural Science Foundation of China under Grant 62402469, U2336206, U2436601, 62121002.

%
%
\bibliographystyle{splncs04}
\bibliography{main}

@String(CVPR= {IEEE Conf. Comput. Vis. Pattern Recog.})

@String(ECCV= {Eur. Conf. Comput. Vis.})

@String(ICASSP=	{ICASSP})

@String(AAAI = {AAAI})

@String(CVPR  = {CVPR})

@String(ECCV  = {ECCV})

@inproceedings{ma2022towards,
  title={Towards Blind Watermarking: Combining Invertible and Non-invertible Mechanisms},
  author={Ma, Rui and Guo, Mengxi and Hou, Yi and Yang, Fan and Li, Yuan and Jia, Huizhu and Xie, Xiaodong},
  booktitle={Proceedings of the 30th ACM International Conference on Multimedia},
  pages={1532--1542},
  year={2022}
}

@article{kirillov2023segment,
  title={Segment anything},
  author={Kirillov, Alexander and Mintun, Eric and Ravi, Nikhila and Mao, Hanzi and Rolland, Chloe and Gustafson, Laura and Xiao, Tete and Whitehead, Spencer and Berg, Alexander C and Lo, Wan-Yen and others},
  journal={arXiv preprint arXiv:2304.02643},
  year={2023}
}

@inproceedings{fernandez2022watermarking,
  title={Watermarking images in self-supervised latent spaces},
  author={Fernandez, Pierre and Sablayrolles, Alexandre and Furon, Teddy and J{\'e}gou, Herv{\'e} and Douze, Matthijs},
  booktitle={ICASSP 2022-2022 IEEE International Conference on Acoustics, Speech and Signal Processing (ICASSP)},
  pages={3054--3058},
  year={2022},
  organization={IEEE}
}

@inproceedings{he2016deep,
  title={Deep residual learning for image recognition},
  author={He, Kaiming and Zhang, Xiangyu and Ren, Shaoqing and Sun, Jian},
  booktitle={Proceedings of the IEEE conference on computer vision and pattern recognition},
  pages={770--778},
  year={2016}
}

@article{paszke2019pytorch,
  title={Pytorch: An imperative style, high-performance deep learning library},
  author={Paszke, Adam and Gross, Sam and Massa, Francisco and Lerer, Adam and Bradbury, James and Chanan, Gregory and Killeen, Trevor and Lin, Zeming and Gimelshein, Natalia and Antiga, Luca and others},
  journal={Advances in neural information processing systems},
  volume={32},
  year={2019}
}

@article{loshchilov2017decoupled,
  title={Decoupled weight decay regularization},
  author={Loshchilov, Ilya and Hutter, Frank},
  journal={arXiv preprint arXiv:1711.05101},
  year={2017}
}

@inproceedings{almohammad2010stego,
  title={Stego image quality and the reliability of PSNR},
  author={Almohammad, Adel and Ghinea, Gheorghita},
  booktitle={2010 2nd International Conference on Image Processing Theory, Tools and Applications},
  pages={215--220},
  year={2010},
  organization={IEEE}
}

@article{wang2004image,
  title={Image quality assessment: from error visibility to structural similarity},
  author={Wang, Zhou and Bovik, Alan C and Sheikh, Hamid R and Simoncelli, Eero P},
  journal={IEEE transactions on image processing},
  volume={13},
  number={4},
  pages={600--612},
  year={2004},
  publisher={IEEE}
}

@inproceedings{wang2024must,
  title={MuST: Robust Image Watermarking for Multi-Source Tracing},
  author={Wang, Guanjie and Ma, Zehua and Liu, Chang and Yang, Xi and Fang, Han and Zhang, Weiming and Yu, Nenghai},
  booktitle={Proceedings of the AAAI Conference on Artificial Intelligence},
  volume={38},
  number={6},
  pages={5364--5371},
  year={2024}
}

@article{bui2023trustmark,
  title={TrustMark: Universal Watermarking for Arbitrary Resolution Images},
  author={Bui, Tu and Agarwal, Shruti and Collomosse, John},
  journal={arXiv preprint arXiv:2311.18297},
  year={2023}
}

@inproceedings{lin2014microsoft,
  title={Microsoft coco: Common objects in context},
  author={Lin, Tsung-Yi and Maire, Michael and Belongie, Serge and Hays, James and Perona, Pietro and Ramanan, Deva and Dollar, Piotr and Zitnick, C Lawrence},
  booktitle={Computer Vision--ECCV 2014: 13th European Conference, Zurich, Switzerland, September 6-12, 2014, Proceedings, Part V 13},
  pages={740--755},
  year={2014},
  organization={Springer}
}

@inproceedings{Agustsson_2017_CVPR_Workshops,
	author = {Agustsson, Eirikur and Timofte, Radu},
	title = {NTIRE 2017 Challenge on Single Image Super-Resolution: Dataset and Study},
	booktitle = {The IEEE Conference on Computer Vision and Pattern Recognition (CVPR) Workshops},
	month = {July},
	year = {2017}
}

@misc{zhang2018unreasonableeffectivenessdeepfeatures,
      title={The Unreasonable Effectiveness of Deep Features as a Perceptual Metric}, 
      author={Richard Zhang and Phillip Isola and Alexei A. Efros and Eli Shechtman and Oliver Wang},
      year={2018},
}

@InProceedings{Lim_2017_CVPR_Workshops,
  author = {Lim, Bee and Son, Sanghyun and Kim, Heewon and Nah, Seungjun and Lee, Kyoung Mu},
  title = {Enhanced Deep Residual Networks for Single Image Super-Resolution},
  booktitle = {The IEEE Conference on Computer Vision and Pattern Recognition (CVPR) Workshops},
  month = {July},
  year = {2017}
}

@inproceedings{fang2022pimog,
  title={PIMoG: An effective screen-shooting noise-layer simulation for deep-learning-based watermarking network},
  author={Fang, Han and Jia, Zhaoyang and Ma, Zehua and Chang, Ee-Chien and Zhang, Weiming},
  booktitle={Proceedings of the 30th ACM international conference on multimedia},
  pages={2267--2275},
  year={2022}
}

@article{wam,
  title={Watermark Anything with Localized Messages},
  author={Sander, Tom and Fernandez, Pierre and Durmus, Alain and Furon, Teddy and Douze, Matthijs},
  journal={arXiv preprint arXiv:2411.07231},
  year={2024}
}

@article{hu2025mask,
  title={Mask Image Watermarking},
  author={Hu, Runyi and Zhang, Jie and Zhao, Shiqian and Lukas, Nils and Li, Jiwei and Guo, Qing and Qiu, Han and Zhang, Tianwei},
  journal={arXiv preprint arXiv:2504.12739},
  year={2025}
}

@article{lu2024robust,
  title={Robust watermarking using generative priors against image editing: From benchmarking to advances},
  author={Lu, Shilin and Zhou, Zihan and Lu, Jiayou and Zhu, Yuanzhi and Kong, Adams Wai-Kin},
  journal={arXiv preprint arXiv:2410.18775},
  year={2024}
}

@misc{blattmann2023stablevideodiffusionscaling,
      title={Stable Video Diffusion: Scaling Latent Video Diffusion Models to Large Datasets}, 
      author={Andreas Blattmann and Tim Dockhorn and Sumith Kulal and Daniel Mendelevitch and Maciej Kilian and Dominik Lorenz and Yam Levi and Zion English and Vikram Voleti and Adam Letts and Varun Jampani and Robin Rombach},
      year={2023},
      eprint={2311.15127},
      archivePrefix={arXiv},
      primaryClass={cs.CV},
}

@article{wan2025,
      title={Wan: Open and Advanced Large-Scale Video Generative Models}, 
      author={Team Wan and Ang Wang and Baole Ai and Bin Wen and Chaojie Mao and Chen-Wei Xie and Di Chen and Feiwu Yu and Haiming Zhao and Jianxiao Yang and Jianyuan Zeng and Jiayu Wang and Jingfeng Zhang and Jingren Zhou and Jinkai Wang and Jixuan Chen and Kai Zhu and Kang Zhao and Keyu Yan and Lianghua Huang and Mengyang Feng and Ningyi Zhang and Pandeng Li and Pingyu Wu and Ruihang Chu and Ruili Feng and Shiwei Zhang and Siyang Sun and Tao Fang and Tianxing Wang and Tianyi Gui and Tingyu Weng and Tong Shen and Wei Lin and Wei Wang and Wei Wang and Wenmeng Zhou and Wente Wang and Wenting Shen and Wenyuan Yu and Xianzhong Shi and Xiaoming Huang and Xin Xu and Yan Kou and Yangyu Lv and Yifei Li and Yijing Liu and Yiming Wang and Yingya Zhang and Yitong Huang and Yong Li and You Wu and Yu Liu and Yulin Pan and Yun Zheng and Yuntao Hong and Yupeng Shi and Yutong Feng and Zeyinzi Jiang and Zhen Han and Zhi-Fan Wu and Ziyu Liu},
      journal = {arXiv preprint arXiv:2503.20314},
      year={2025}
}

@article{kong2024hunyuanvideo,
  title={Hunyuanvideo: A systematic framework for large video generative models},
  author={Kong, Weijie and Tian, Qi and Zhang, Zijian and Min, Rox and Dai, Zuozhuo and Zhou, Jin and Xiong, Jiangfeng and Li, Xin and Wu, Bo and Zhang, Jianwei and others},
  journal={arXiv preprint arXiv:2412.03603},
  year={2024}
}

@inproceedings{hu2025robust,
  title={Robust-wide: Robust watermarking against instruction-driven image editing},
  author={Hu, Runyi and Zhang, Jie and Xu, Ting and Li, Jiwei and Zhang, Tianwei},
  booktitle={European Conference on Computer Vision},
  pages={20--37},
  year={2025},
  organization={Springer}
}

@article{fernandez2024video,
  title={Video Seal: Open and Efficient Video Watermarking},
  author={Fernandez, Pierre and Elsahar, Hady and Yalniz, I. Zeki and Mourachko, Alexandre},
  journal={arXiv preprint arXiv:2412.09492},
  year={2024}
}

@misc{ye2023itovefficientlyadaptingdeep,
      title={ItoV: Efficiently Adapting Deep Learning-based Image Watermarking to Video Watermarking}, 
      author={Guanhui Ye and Jiashi Gao and Yuchen Wang and Liyan Song and Xuetao Wei},
      year={2023},
      eprint={2305.02781},
      archivePrefix={arXiv},
      primaryClass={cs.CR},
}

@article{yang2024cogvideox,
  title={CogVideoX: Text-to-Video Diffusion Models with An Expert Transformer},
  author={Yang, Zhuoyi and Teng, Jiayan and Zheng, Wendi and Ding, Ming and Huang, Shiyu and Xu, Jiazheng and Yang, Yuanming and Hong, Wenyi and Zhang, Xiaohan and Feng, Guanyu and others},
  journal={arXiv preprint arXiv:2408.06072},
  year={2024}
}

@inproceedings{huang2018munit,
  title={Multimodal Unsupervised Image-to-image Translation},
  author={Huang, Xun and Liu, Ming-Yu and Belongie, Serge and Kautz, Jan},
  booktitle={ECCV},
  year={2018}
}

@ARTICLE{475889,
  author={Chun-Hsien Chou and Yun-Chin Li},
  journal={IEEE Transactions on Circuits and Systems for Video Technology}, 
  title={A perceptually tuned subband image coder based on the measure of just-noticeable-distortion profile}, 
  year={1995},
  volume={5},
  number={6},
  pages={467-476},
  }

@Article{liu2022convnet,
  author  = {Zhuang Liu and Hanzi Mao and Chao-Yuan Wu and Christoph Feichtenhofer and Trevor Darrell and Saining Xie},
  title   = {A ConvNet for the 2020s},
  journal = {Proceedings of the IEEE/CVF Conference on Computer Vision and Pattern Recognition (CVPR)},
  year    = {2022},
}

@misc{rombach2021highresolution,
      title={High-Resolution Image Synthesis with Latent Diffusion Models}, 
      author={Robin Rombach and Andreas Blattmann and Dominik Lorenz and Patrick Esser and Björn Ommer},
      year={2021},
      eprint={2112.10752},
      archivePrefix={arXiv},
      primaryClass={cs.CV}
}

@article{Qwen2.5-VL,
  title={Qwen2.5-VL Technical Report},
  author={Bai, Shuai and Chen, Keqin and Liu, Xuejing and Wang, Jialin and Ge, Wenbin and Song, Sibo and Dang, Kai and Wang, Peng and Wang, Shijie and Tang, Jun and Zhong, Humen and Zhu, Yuanzhi and Yang, Mingkun and Li, Zhaohai and Wan, Jianqiang and Wang, Pengfei and Ding, Wei and Fu, Zheren and Xu, Yiheng and Ye, Jiabo and Zhang, Xi and Xie, Tianbao and Cheng, Zesen and Zhang, Hang and Yang, Zhibo and Xu, Haiyang and Lin, Junyang},
  journal={arXiv preprint arXiv:2502.13923},
  year={2025}
}

@article{videoworldsimulators2024,
  title={Video generation models as world simulators},
  author={Tim Brooks and Bill Peebles and Connor Holmes and Will DePue and Yufei Guo and Li Jing and David Schnurr and Joe Taylor and Troy Luhman and Eric Luhman and Clarence Ng and Ricky Wang and Aditya Ramesh},
  year={2024},
  url={https://openai.com/research/video-generation-models-as-world-simulators},
}

@article{kling,
  title={Kling},
  author={Kuaishou},
  year={2024},
  url={https://kling.kuaishou.com/},
}

@article{hailuo,
  title={Hailuo ai Video Generator},
  author={Reimagine Video Creation },
  year={2024},
  url={https://hailuoai.video/},
}

@inproceedings{teed2020raft,
  title={Raft: Recurrent all-pairs field transforms for optical flow},
  author={Teed, Zachary and Deng, Jia},
  booktitle={European conference on computer vision},
  pages={402--419},
  year={2020},
  organization={Springer}
}

@InProceedings{huang2023vbench,
     title={{VBench}: Comprehensive Benchmark Suite for Video Generative Models},
     author={Huang, Ziqi and He, Yinan and Yu, Jiashuo and Zhang, Fan and Si, Chenyang and Jiang, Yuming and Zhang, Yuanhan and Wu, Tianxing and Jin, Qingyang and Chanpaisit, Nattapol and Wang, Yaohui and Chen, Xinyuan and Wang, Limin and Lin, Dahua and Qiao, Yu and Liu, Ziwei},
     booktitle={Proceedings of the IEEE/CVF Conference on Computer Vision and Pattern Recognition},
     year={2024}
 }

@article{su2025safe,
  title={Safe-Sora: Safe Text-to-Video Generation via Graphical Watermarking},
  author={Su, Zihan and Qiu, Xuerui and Xu, Hongbin and Jiang, Tangyu and Zhuang, Junhao and Yuan, Chun and Li, Ming and He, Shengfeng and Yu, Fei Richard},
  journal={arXiv preprint arXiv:2505.12667},
  year={2025}
}

@INPROCEEDINGS{10888991,
  author={Liu, Xiaohang and Chang, Heng and Wei, Jinfu and Zhu, Lei and Liu, Li and Li, Likun and Zhou, Shiji and Li, Chengyuan and Xu, Di and Gao, Wei},
  booktitle={ICASSP 2025 - 2025 IEEE International Conference on Acoustics, Speech and Signal Processing (ICASSP)}, 
  title={Implanting Robust Watermarks in Latent Diffusion Models for Video Generation}, 
  year={2025},
  volume={},
  number={},
  pages={1-5},
  doi={10.1109/ICASSP49660.2025.10888991}}

@ARTICLE{11299284,
  author={Yang, Jin and Zhang, Wenzhu and Yang, Linlin and Zeng, Shunyang and Gao, Tianyu},
  journal={IEEE Transactions on Circuits and Systems for Video Technology}, 
  title={RINGet: A Robust Watermarking Framework for Diffusion-Based Video Generation}, 
  year={2026},
  volume={36},
  number={5},
  pages={7184-7195},
  doi={10.1109/TCSVT.2025.3643532}}
\end{document}